\documentclass[letterpaper, 10 pt, conference]{ieeeconf}
\IEEEoverridecommandlockouts
\usepackage{graphics}
\usepackage{hyperref}
\usepackage{epsfig}
\usepackage{mathptmx}
\usepackage{times}
\usepackage{amsmath}
\usepackage{amssymb}
\usepackage{cite}
\usepackage{booktabs}
\usepackage{tabularx}
\usepackage{multirow}
\usepackage{ragged2e}
\usepackage{subcaption}
\usepackage[table]{xcolor}
\definecolor{settingblue}{HTML}{EAF2F8}
\newcommand{\settingrow}{\rowcolor{settingblue}}

\newif\ifshowtodo \showtodofalse

\newcommand{\chk}[1]{\ifshowtodo{\color{orange}#1}\else#1\fi}
\definecolor{oneround}{RGB}{176,92,0}
\newif\ifsingle \singlefalse

\newcommand{\vlm}{Intern-S2-Preview-397B}
\newcommand{\tabnote}[1]{%
  \par\vspace{3pt}%
  {\fontsize{7.5}{8.7}\selectfont
   \justifying
   \noindent #1\par}%
}

\title{\LARGE \bf
MemTransfer: Benchmarking Memory Beyond Matched Experience in Embodied Decision-Making
}
\author{
    Haiming Tang$^{1*}$, 
    Xianjie Dai$^{2*}$, 
    Gujie Shao$^{1}$, 
    Zuyi Guo$^{2}$, \\
    Jingguang Li$^{2}$, 
    Kailang Ma$^{2}$, 
    Yihong Tang$^{3}$, 
    Heye Huang$^{2\dagger}$%
    \thanks{*Equal contribution.}%
    \thanks{$^{\dagger}$Corresponding author: {\tt\small heye.huang@kaist.ac.kr}.}%
    \thanks{This work was conducted as part of the MIT-UF-NEU Joint Summer Research Camp 2026.}%
    \\[1.5ex]
    $^{1}$National University of Singapore \\
    $^{2}$Korea Advanced Institute of Science and Technology  \\
    $^{3}$McGill University
}

\begin{document}
\bstctlcite{IEEEexample:BSTcontrol}

\maketitle
\thispagestyle{empty}
\pagestyle{empty}

\AddToHookNext{env/abstract/end}{%
  \ifhmode\unskip\space\fi
  Code is available at \url{https://github.com/PKUTHM/MemTransfer_public}.%
}
\begin{abstract}
Memory lets an embodied agent reuse past experience, yet retaining useful information does not ensure that the agent can apply it when conditions change. We present MemTransfer, a benchmark comparing six memory representations, a working-memory baseline and five representations of past experience, under a shared frozen vision-language-model policy. It comprises 100 navigation cases across ten task types in a simulated warehouse, with expert demonstrations supplying the history. Three comparisons vary the starting pose, route availability, and amount and task relevance of history. With one demonstration per task, Full-context and Episodic memory reach 95.3\% and 100.0\% success at the original demonstration start, but lose 48–49 percentage points at a new test start. Summary changes little between these two test starts, yet with four demonstrations per task it retains a smaller fraction of its unchanged-route success after blocking (39.3\%) than Working memory (44.8\%) or the two trajectory memories (56–58\%). At the new test start, increasing from one to four relevant demonstrations raises Episodic success by 14.3 percentage points, while the other evaluated representations gain no more than 1.3 percentage points. Replacing half of the relevant histories with other-task experience lowers success for both trajectory memories. These results show that robustness to one kind of mismatch does not imply robustness to another, motivating evaluation of both stored information and its use at decision time. 

\end{abstract}

\section{INTRODUCTION}
\label{sec:intro}

\begin{figure*}[t]
\centering
\includegraphics[width=\textwidth]{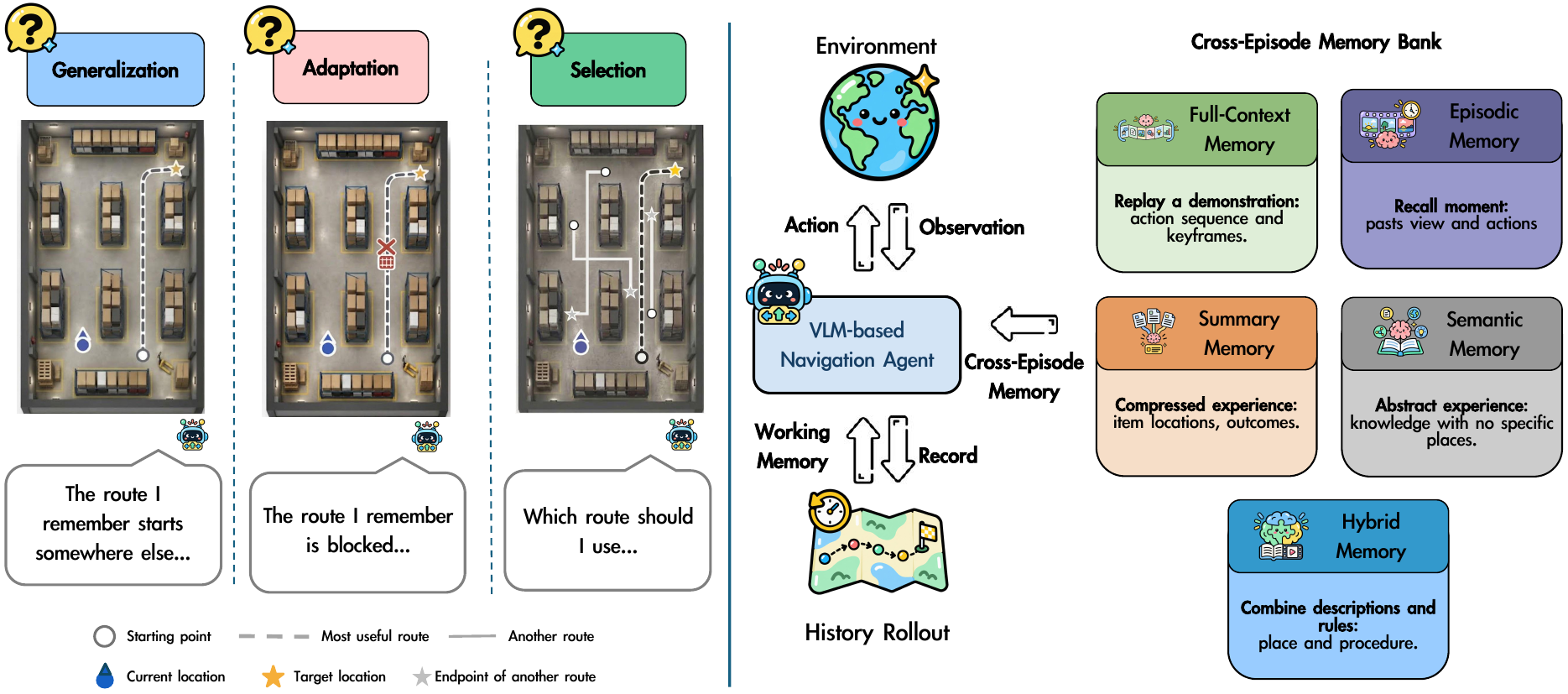}

\caption{\textbf{Three questions about reusing past experience.} \emph{Left:} a new starting pose (RQ1), a blocked route (RQ2), and multiple histories including experience from other tasks (RQ3). \emph{Right:} a shared policy and working-memory window, with five representations of past experience.}
\label{fig:teaser}
\end{figure*}

A robot working repeatedly in the same warehouse can use experience from earlier tasks to guide later ones. If it has already found an object, that experience may save it from searching again~\cite{goatbench24cvpr}. A memory representation determines what information is kept from the earlier task and how the policy receives it when deciding what to do next.
The next task may begin elsewhere in the warehouse, requiring the robot to use its experience from a new starting point. Even a familiar route can become unusable if an aisle is blocked. Over time, the robot will also accumulate histories collected for different goals, raising the question of which experience is useful for the task at hand.

Success when repeating a task under the original conditions does not tell us how well memory will help in these situations. Different representations may respond differently to the same change: preserving useful information does not necessarily make it easy to apply in a new episode. We therefore compare six memory representations under a common policy, asking how their benefits depend on the experience available and the conditions in which it is used. Three research questions guide the comparison.

\begin{samepage}
\noindent\textbf{RQ1 --- Generalization.} How does a mismatch in starting position or orientation between past experience and the current episode affect the performance of different memory representations?\par
\end{samepage}

\begin{samepage}
\noindent\textbf{RQ2 --- Adaptation.} How do different memory representations support task completion and recovery when an environmental change makes past experience partially outdated?\par
\end{samepage}

\begin{samepage}
\noindent\textbf{RQ3 --- Using multiple histories.} How do the amount and task relevance of past experience affect the performance of different memory representations?\par
\end{samepage}

To study these questions, we introduce MemTransfer, a collection of 100 navigation task cases across ten task types in a simulated warehouse. Expert trajectories provide the historical experience from which memory is constructed. We compare a working-memory baseline and five representations of past experience using the same frozen Vision-Language-Model (VLM) policy and interaction interface.


The results show that Full-context and Episodic memory can achieve high success when the test start matches the demonstration, yet decline substantially at a new start. Summary changes less across start poses, but this stability does not carry over to blocked routes: it retains a smaller fraction of its unchanged-route success than Working memory, Full-context or Episodic. Additional relevant histories benefit Episodic memory, while replacing relevant histories with other-task experience reduces success for both Episodic and Full-context. These findings highlight the importance of testing how memory is reused across conditions.


In summary, our main contributions are as follows:
\begin{itemize}
    \item We present a controlled benchmark with 100 navigation cases across ten task types and paired expert trajectories, designed to study memory reuse under changes in start pose, route availability, and prior experience.

    \item We establish a unified evaluation protocol for six memory representations and controlled comparisons of memory generalization, adaptation, and multi-history use.

    \item We demonstrate that memory representations exhibit distinct trade-offs: trajectory-based memories excel under matched conditions but degrade under mismatch, while abstracted memories transfer more consistently but remain sensitive to outdated experience.
\end{itemize}
\section{RELATED WORK}

\textbf{Embodied benchmarks.} Simulation suites such as Habitat~\cite{habitat19iccv}, Gibson~\cite{gibson18cvpr}, ALFRED~\cite{alfred20cvpr} and BEHAVIOR~\cite{behavior21corl} made embodied evaluation standard, and EmbodiedBench~\cite{embodiedbench25icml} and OmniNavBench~\cite{omninavbench26arxiv} widen the range of skills and robot bodies. All share one assumption: every trial is a fresh episode, with the environment reset and no history kept across runs. Three lines of work let memory persist: iterative VLN evaluates agents over tours of up to 100 instructions in one scene and finds that only map-building agents profit~\cite{ivln23cvpr}, GOAT-Bench evaluates lifelong navigation to sequences of goals~\cite{goatbench24cvpr}, and MultiON tests map memory within an episode~\cite{multion20neurips}. However, none of these methods controls the relation between the earlier trajectories and the tested episode.

\textbf{Memory representations.} Seen through what reaches the policy at decision time, existing designs fall into a few classes. A \emph{window} over recent steps is the working memory of language agents~\cite{memgpt23arxiv} and the keyframe buffer of RoboMemArena~\cite{robomemarena26arxiv}. Summaries and knowledge distilled from experience appear in the running summaries and stored procedures of generative agents~\cite{generativeagents23uist,evo_memory25arxiv}. \emph{Retrieval} of the past moment that matches the present runs from semi-parametric topological memory~\cite{sptm18iclr} and scene memory transformers~\cite{smt19cvpr} to the buffer-plus-store designs of VLM$^2$~\cite{vlm2_25arxiv} and 3DLLM-Mem~\cite{3dllm_mem25nips}, and to ReMEmbR~\cite{remembr25icra} and Embodied-RAG~\cite{embodiedrag24arxiv}, which query hours of traversal. Providing a recorded trajectory as \emph{Full-context} memory supplies in-context guidance. We compare these five memory representations and their Summary--Semantic combination, Hybrid, each realised as a prompt the same frozen VLM can read. Metric and topological maps~\cite{conceptgraphs24icra,topological20cvpr,dynamem25icra} are the one family we do not include as a representation: they need a map builder and a planner outside the policy, so a comparison through one fixed prompt cannot separate the memory from that machinery. This exclusion is limited to explicit map structures, while both Summary and Semantic retain spatial and navigational knowledge.

\textbf{Memory benchmarks.} LongBench~\cite{longbench24acl} tests recall from a long context; MemoryAgentBench~\cite{memoryagentbench25iclr} and EgoMemReason~\cite{egomemreason26arxiv} test multi-session retrieval, test-time learning and reasoning over egocentric video. Closer to robotics, FindingDory~\cite{findingdory25iclr} navigates from a day of the agent's own interaction in Habitat, and WorldLines~\cite{worldlines26arxiv} tracks changing household states over multi-day traces.

MemTransfer complements these benchmarks by varying the relation between past experience and the current episode: mismatched starts, outdated routes, and changes in the amount and task relevance of history.

\section{THE MemTransfer BENCHMARK}
\label{sec:benchmark}

\begin{figure*}[t]
\centering
\includegraphics[width=\textwidth]{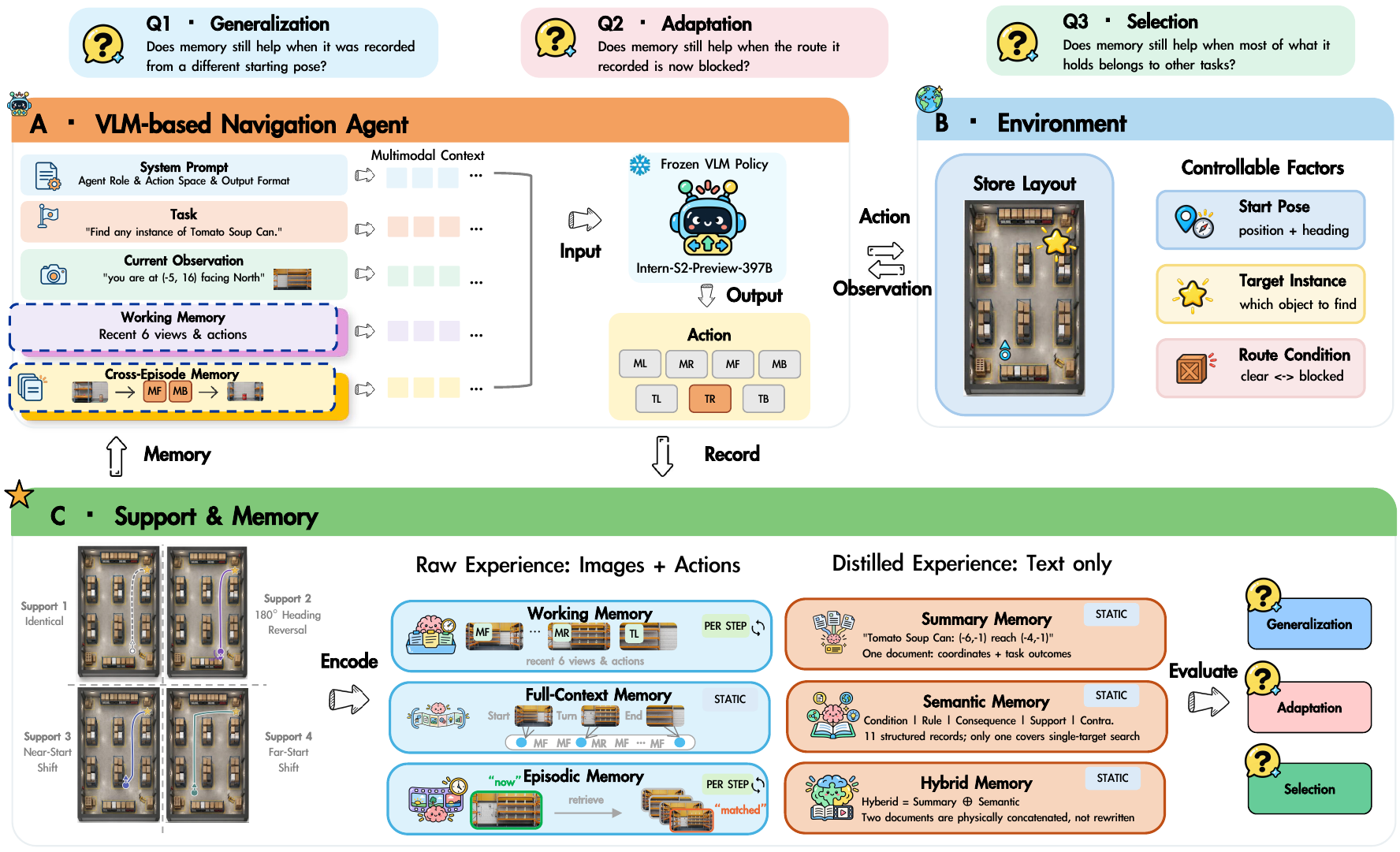}
\caption{\textbf{Overview.} \textbf{(A)}~The shared VLM policy receives the task, observation, pose, action feedback, recent interactions and historical memory. \textbf{(B)}~Warehouse layout, start poses, target instances and route obstructions. \textbf{(C)}~Demonstration starting poses and the six memory representations, with examples of their contents and input sizes. Section~\ref{sec:variants} defines the comparisons used for each RQ.}
\label{fig:method}
\end{figure*}

MemTransfer uses warehouse navigation to study how a fixed policy uses past experience. We pair each task with expert demonstrations and test episodes, varying the starting pose (RQ1), route availability (RQ2), and the amount and relevance of history (RQ3). Figure~\ref{fig:method} shows the environment and the shared interface used to compare memory representations.

\subsection{Environment and Tasks}
\label{sec:env}

\textbf{Scene.} We use an industrial warehouse built in NVIDIA Isaac Sim~4.5, covering approximately $16\,\text{m}\times 30\,\text{m}$ (Fig.~\ref{fig:method}B). Six rack blocks form two banks connected by a cross-aisle, with three parallel corridors in each bank and two additional racks at the north and south ends. The layout provides alternative routes between locations, allowing us to block a familiar passage while keeping the task solvable in RQ2. Items in the same category are grouped in neighbouring bays, so a previously observed location can provide a useful cue for later search.

\textbf{Interface.} The floor is divided into $1\,\text{m}$ cells, and the agent faces one of four cardinal directions. The policy receives the task instruction, its absolute position and heading, an egocentric RGB image, and action feedback. The camera is $1.2\,\text{m}$ above the floor with a $90^\circ$ field of view. The environment supplies neither a map nor an object-location list; knowledge of object locations comes from observations or memory. The agent can move one cell forward, backward, left or right without changing heading, or turn left, right or around. Racks and walls are not traversable: a blocked move leaves the agent in place, returns feedback, and consumes one step. An item is reached from the cell directly in front of its rack face. Episodes end at task completion or the step limit $T_{\max}$. All representations use this interface; the memory information supplied to the policy is specified in Sec.~\ref{sec:representations}.

\textbf{Tasks.} We define 10 navigation task types and instantiate 10 cases of each, for 100 cases in total (Table~\ref{tab:task_taxonomy}). Each case specifies the target items or category and an original starting position and heading. The tasks cover \emph{basic search}, \emph{multi-object search}, and \emph{loops} that return to the current episode's origin or revisit the same object instance. Past experience may provide target locations and route guidance, while the agent must also track progress in the current episode. The three RQs use these same task cases with the conditions defined in Sec.~\ref{sec:variants}; Sec.~\ref{sec:tasktype} further examines how memory benefits vary across task groups.

\begin{table}[t]
\centering
\caption{\textbf{Navigation tasks used in the three RQs.} }
\label{tab:task_taxonomy}
\setlength{\tabcolsep}{4pt}
\fontsize{7.5pt}{9pt}\selectfont
\begin{tabularx}{\columnwidth}{@{}l
  >{\hsize=1.2\hsize\linewidth=\hsize
    \raggedright\arraybackslash}X
  >{\hsize=0.8\hsize\linewidth=\hsize
    \raggedright\arraybackslash}X@{}}
\toprule
\textbf{Type} & \textbf{Task and route} & \textbf{Task requirement} \\
\midrule
\multicolumn{3}{l}{\textit{Basic search}} \\
1 & Single object search, $O \to A$ & Reach one instance of $A$ \\
2 & Category object search, $O \to C$ & Reach one item in $C$ \\
\midrule
\multicolumn{3}{l}{\textit{Multi-object search}} \\
3 & Exhaustive single object search, $O \to A_1 \to \dots \to A_k$ & Visit every instance of $A$ \\
4 & Exhaustive category search, $O \to C_1 \to \dots \to C_m$ & Visit every item in $C$ \\
5 & Unordered search, $O \to \{A, B\}$ & Visit both target items \\
6 & Ordered search, $O \to A \to B$ & Visit $A$ before $B$ \\
\midrule
\multicolumn{3}{l}{\textit{Loop}} \\
7 & Search with return, $O \to A \to O$ & Return to the episode origin \\
8 & Exhaustive search with return, $O \to \{A_k\} \to O$ & Visit all, then return \\
9 & Origin--A--B--A, $O \to A_i \to B \to A_i$ & Return to the same $A$ \\
10 & Origin--A--B--origin, $O \to A \to B \to O$ & Visit $A$, then $B$, then return \\
\bottomrule
\end{tabularx}
\tabnote{Each type has 10 cases. $O$: episode origin; $A$, $B$: target object types; $C$: category. Subscripts identify instances; braces indicate visits without a prescribed order.}
\end{table}

\subsection{Evaluation Design}
\label{sec:variants}

For each of the 100 task cases, we collect four successful demonstrations: one from the \emph{original start}, one from the same position with the opposite heading, and two from nearby and distant starts. We also prepare a test episode from an initial pose different from those of the four demonstrations for that case, together with a copy in which a route is blocked. The four demonstrations are used to construct memory; the two test reference trajectories are excluded from memory construction.

Table~\ref{tab:main} distinguishes the \emph{test start}, \emph{route condition}, and \emph{number of demonstrations per task}. One demonstration means the original trajectory; four demonstrations add the other three starting poses. These counts describe the available experience per task, not the number of trajectories in every memory: text representations consolidate the selected demonstrations across all 100 task cases. We use these settings in three comparisons.

\textbf{RQ1: Generalization across start poses.} First, we build memory from the original demonstration and compare the agent's performance from the \emph{original start} and the \emph{new start} (Table~\ref{tab:main}). Second, we keep the test episode at the original start and change which single demonstration builds memory: original start, \emph{opposite heading}, \emph{nearby start}, or \emph{distant start} (Table~\ref{tab:transfer}). The first comparison changes the test start; the second changes the start of the remembered experience. Both use the unchanged warehouse.

\textbf{RQ2: Acting with outdated memory.} We keep the new test start and the four demonstrations fixed, then block a passage shared by the test reference route and at least one demonstration. An alternative route remains available in every case. Comparing the unchanged and blocked route conditions examines task completion and recovery when part of the stored route information becomes outdated (Tables~\ref{tab:main} and~\ref{tab:rq2}).

\textbf{RQ3: Using multiple histories.} With the new test start and warehouse unchanged, we first compare one versus four demonstrations per task to examine the benefit of additional relevant experience (Table~\ref{tab:main}). We then keep four demonstrations but replace two with histories from other tasks with different target requirements (Fig.~\ref{fig:rq3}). These histories either share no poses with the test reference route (\emph{irrelevant}), or share poses and are selected for potentially conflicting continuations (\emph{confusable}); a pose includes position and heading. Replacement removes relevant experience as well as adding other tasks' histories.

\textbf{Trajectory construction.} All 600 reference trajectories are near-optimal planner solutions on the ground-truth map, verified by replay. The opposite-heading demonstration turns $180^\circ$ at the original cell. Nearby starts move 2--4 cells along the same aisle, retain the heading, and share at least half the original route's edges. Distant starts move at least six cells to another aisle, turn $90^\circ$, and share at most half its edges. The new test start keeps the target out of the first view and has a reference length within $0.75$--$1.35$ of the original demonstration length.

\subsection{Memory Representations}
\label{sec:representations}

We compare a working-memory baseline with five representations of past demonstrations under the same frozen VLM policy, \vlm{}~\cite{bai2026intern}. Table~\ref{tab:memory_representations} summarizes what each stores and supplies at decision time. We distinguish memory construction from access~\cite{remembr25icra,embodiedrag24arxiv,evo_memory25arxiv}:
\begin{equation}
\begin{aligned}
    M_r &= B_r(D_r),\\
    m_{r,t} &= A_r(M_r,x_t),\\
    a_t &\sim \pi_\theta(\cdot\mid x_t,W_t,m_{r,t}),
\end{aligned}
\label{eq:memory_policy}
\end{equation}
where $B_r$ builds memory $M_r$ from the demonstrations $D_r$ specified by the evaluation condition (Sec.~\ref{sec:variants}), and $A_r$ selects the information $m_{r,t}$ supplied at step $t$. The current input $x_t$ contains the task, image $o_t$, pose and action feedback. All configurations share the recent-interaction window $W_t$; the baseline has no historical memory, $M_r=m_{r,t}=\varnothing$.

\textbf{Storage and use during testing.} The stored memory $M_r$ and policy parameters $\theta$ remain fixed, while $W_t$ updates as the episode proceeds. Full-context, Summary, Semantic and Hybrid supply fixed historical content at each step. Episodic memory instead retrieves a local segment using the current view. RQ2 therefore measures task completion and recovery with fixed historical memory.

\begin{table*}[t]
\centering
\caption{\textbf{Construction and use of the six memory representations.} Each historical representation supplements the working-memory window used by the baseline.}
\label{tab:memory_representations}
\small
\setlength{\tabcolsep}{5pt}
\renewcommand{\arraystretch}{1.15}
\begin{tabularx}{\textwidth}{@{}
>{\raggedright\arraybackslash}p{0.13\textwidth}
>{\hsize=1.05\hsize\linewidth=\hsize\raggedright\arraybackslash}X
>{\hsize=1\hsize\linewidth=\hsize\raggedright\arraybackslash}X
>{\hsize=0.95\hsize\linewidth=\hsize\raggedright\arraybackslash}X
@{}}
\toprule
\textbf{Representation} & \textbf{Memory organization} & \textbf{Construction} & \textbf{Decision-time access} \\
\midrule
Working memory & Ordered recent interactions & Keep the last $h$ steps & Entire recent window \\
\midrule
Full-context & Chronological actions and keyframes & Retain actions; select keyframes & All retained actions and keyframes \\
Summary & Object locations, landmarks and routes & Sequential summarization & Entire summary document \\
Semantic & Generalized spatial and navigation knowledge & Extract, consolidate and refine knowledge from demonstrations & Entire knowledge document \\
Hybrid & Summary and Semantic documents & Concatenate the two documents & Both documents \\
Episodic & Frames linked to expert continuations & Encode frames; retain trajectories & View-matched local continuation \\
\bottomrule
\end{tabularx}
\end{table*}

Full-context keyframes cover the start, sub-goal completions, end and moments before turns, supplemented by evenly sampled frames. Image budgets and window lengths are given in Sec.~\ref{sec:setup}.

\textbf{Episodic retrieval.} Let $I$ index the stored demonstration frames $o_u$, $\phi$ be the ViT-B/16 encoder, and $d$ the matcher's embedding distance. Retrieval matches the current view to a stored frame and returns its continuation:
\begin{equation}
\begin{aligned}
    u_t &\in \underset{u\in I}{\arg\min}\;d\!\left(\phi(o_t),\phi(o_u)\right),\\
    m_{\mathrm{epi},t} &= \operatorname{Segment}_{H_{\mathrm{epi}}}(D_{\mathrm{epi}},u_t).
\end{aligned}
\label{eq:episodic_read}
\end{equation}
The segment includes the matched view and up to $H_{\mathrm{epi}}$ subsequent expert actions with their resulting views, stopping at the end of the demonstration. Ties are broken by task stage and elapsed steps. The VLM uses this context to choose its next action.

\section{EXPERIMENTS}
\label{sec:experiments}


We examine start-pose generalization (RQ1), outdated memory (RQ2), and multiple histories (RQ3). Tables~\ref{tab:main}--\ref{tab:rq2} and Fig.~\ref{fig:rq3} summarize the main evaluated conditions. Results are then reported by task group, followed by key design choices and limitations.

\subsection{Experimental Setup}
\label{sec:setup}

We use the same frozen \vlm{} at temperature $0$, with a $30$-token answer budget and reasoning disabled. Episodes end upon task completion or after $T_{\max}=200$ steps, with at most $50$ images per prompt. Working memory keeps $h=6$ steps. Full-context memory uses up to $40$ reference images, while Episodic memory returns $H_{\mathrm{epi}}=6$ actions and their resulting views, for at most $7$ reference images. These settings remain fixed unless otherwise stated.

Each configuration is evaluated on the same $100$ cases. Results for the new test start, blocked routes, changes to the demonstration start, and mixed histories use three runs; the original-start design-choice ablations use five. Summary and Semantic documents are constructed once from the selected demonstrations across all $100$ cases and remain unchanged during testing. Hybrid coverage is described in Sec.~\ref{sec:limitations}.

\subsection{Evaluation Metrics}
\label{sec:metrics}

For each run of $N=100$ cases, let $s_i=1$ if case $i$ visits all required standpoints, respects any prescribed order, and ends at the required location within $T_{\max}$; otherwise $s_i=0$. Let $p_i$ be the executed step count and $l_i$ the reference expert length for the test episode. Success Rate (SR) is expressed as a percentage; failures contribute zero to Success weighted by Path Length (SPL). We report
\begin{equation}
\begin{aligned}
    \mathrm{SR} &= \frac{100}{N}\sum_{i=1}^{N}s_i,\\
    \mathrm{SPL} &= \frac{1}{N}\sum_{i=1}^{N}s_i\frac{l_i}{\max(p_i,l_i)}.
\end{aligned}
\label{eq:sr_spl}
\end{equation}

For RQ2, let $S_{r,0}$ and $S_{r,1}$ denote the reported mean SR for representation $r$ under the unchanged and blocked route conditions, and let $w$ denote Working memory alone. Table~\ref{tab:rq2} reports the baseline-relative change $\Delta_r$ and retention $R_r$:
\begin{equation}
\begin{aligned}
    \Delta_r &= (S_{r,1}-S_{r,0})-(S_{w,1}-S_{w,0}),\\
    R_r &= S_{r,1}/S_{r,0}.
\end{aligned}
\label{eq:change_retention}
\end{equation}
A negative $\Delta_r$ indicates a larger SR drop than the working-memory baseline, in percentage points. Retention is the fraction of unchanged-route SR retained after blocking.

For blocked episodes, let $b_i=1$ if the agent attempts to enter a newly blocked cell at least once, and $0$ otherwise. Follow rate $F$ and recovery rate $C$ are
\begin{equation}
    F=\frac{100}{N}\sum_{i=1}^{N}b_i,\qquad
    C=100\frac{\sum_{i=1}^{N}b_i s_i}{\sum_{i=1}^{N}b_i}.
\label{eq:follow_recovery}
\end{equation}
Both are percentages; recovery measures success among episodes that attempted a blocked cell. SR, SPL, follow and recovery are computed per run; reported $\pm$ values indicate standard deviations across runs. We also report images per request and total prompt-plus-completion tokens per $100$ cases.


\begin{table*}[t]
\centering
\caption{\textbf{Memory performance across the three RQs.}}
\label{tab:main}
\scriptsize
\setlength{\tabcolsep}{2.3pt}
\begin{tabularx}{\textwidth}{@{}lcc*{8}{>{\centering\arraybackslash}X}@{}}
\toprule
\multicolumn{3}{l}{\textbf{Test start}} & \multicolumn{2}{c}{Original} & \multicolumn{2}{c}{New} & \multicolumn{2}{c}{New} & \multicolumn{2}{c}{New} \\
\multicolumn{3}{l}{\textbf{Route condition}} & \multicolumn{2}{c}{Unchanged} & \multicolumn{2}{c}{Unchanged} & \multicolumn{2}{c}{Unchanged} & \multicolumn{2}{c}{Blocked} \\
\multicolumn{3}{l}{\textbf{Demonstrations per task}} & \multicolumn{2}{c}{1} & \multicolumn{2}{c}{1} & \multicolumn{2}{c}{4} & \multicolumn{2}{c}{4} \\
\cmidrule(lr){4-5} \cmidrule(lr){6-7} \cmidrule(lr){8-9} \cmidrule(lr){10-11}
\textbf{Representation} & \textbf{Img} & \textbf{Mtok} & SR\,$\uparrow$ & SPL\,$\uparrow$ & SR\,$\uparrow$ & SPL\,$\uparrow$ & SR\,$\uparrow$ & SPL\,$\uparrow$ & SR\,$\uparrow$ & SPL\,$\uparrow$ \\
\midrule
Working memory & $7$  & $34$ & \chk{$15.2 \pm 3.0$} & \chk{$0.023$} & 
\chk{$22.3 \pm 2.9$} & \chk{$0.051$} & \chk{$22.3 \pm 2.9$} & \chk{$0.051$}
& $10.0 \pm 2.6$ & $0.025$ \\
\midrule
Full-context        & $47$ & $96$ / $138$ & $95.3 \pm 0.6$ & $0.792$ & $46.7 \pm 2.5$ & $0.157$ & $48.0 \pm 2.6$ & $0.167$ & $28.0 \pm 3.5$ & $0.067$ \\
Summary             & $7$  & $163$ / $588$ & $43.0 \pm 2.7$ & $0.136$ & $44.3 \pm 3.8$ & $0.149$ & $44.0 \pm 2.6$ & $0.173$ & $17.3 \pm 3.1$ & $0.042$ \\
Semantic            & $7$  & $80$ / $233$  & $15.0 \pm 2.5$ & $0.050$ & $19.7 \pm 3.5$ & $0.063$ & $17.0 \pm 1.7$ & $0.061$ & $4.3 \pm 1.5$ & $0.007$ \\
Hybrid              & $7$  & $216$ / ---   & $39.0 \pm 3.0$ & $0.128$ & $38.0 \pm 1.0$ & $0.125$ & --- & --- & --- & --- \\
\settingrow
Episodic            & $14$ & $43$ / $\mathbf{35}$ & $\mathbf{100.0 \pm 0.0}$ & $\mathbf{0.849}$ & $\mathbf{51.7 \pm 3.5}$ & $\mathbf{0.217}$ & $\mathbf{66.0 \pm 1.0}$ & $\mathbf{0.295}$ & $\mathbf{37.0 \pm 3.0}$ & $\mathbf{0.106}$ \\
\bottomrule
\end{tabularx}
\tabnote{Settings follow Sec.~\ref{sec:variants}. Each run uses 100 cases. SR [\%]: mean $\pm$ std; SPL: mean. Working memory uses no demonstrations; its new-start baseline is shared. \emph{Img}: images per request. \emph{Mtok}: million prompt$+$completion tokens per 100 new-start cases, with 1 / 4 demonstrations per task. The original-start column is a replay reference: the test episode starts from the
pose its demonstration was recorded from. Unreported Hybrid settings: Sec.~\ref{sec:limitations}.}
\end{table*}

Table~\ref{tab:main} reports SR, SPL and inference cost for the six representations under the tested conditions. The comparison between original and new test starts addresses RQ1, the comparison between unchanged and blocked routes addresses RQ2, and the comparison of one versus four demonstrations per task addresses RQ3. The following sections interpret these results alongside the corresponding auxiliary tables.

\subsection{RQ1: Generalization Across Start Poses}
\label{sec:rq1}

\begin{table}[t]
\centering
\caption{\textbf{RQ1: Changing starting pose.}}
\label{tab:transfer}
\footnotesize
\setlength{\tabcolsep}{3pt}
\begin{tabular*}{\columnwidth}{@{\extracolsep{\fill}}lcccc@{}}
\toprule
\multicolumn{5}{l}{\scriptsize\emph{Test: original start; memory: one demonstration per task}} \\ \cmidrule(lr){2-5} \textbf{Representation} & \shortstack{\emph{Original}\\\emph{start}} & \shortstack{\emph{Opposite}\\\emph{heading}} & \shortstack{\emph{Nearby}\\\emph{start}} & \shortstack{\emph{Distant}\\\emph{start}} \\
\midrule
Full-context        & $95.3 \pm 0.6$ & $83.0 \pm 3.0$ & $77.7 \pm 3.1$ & $33.7 \pm 4.7$ \\
                    & \footnotesize$0.792$ & \footnotesize$0.487$ & \footnotesize$0.487$ & \footnotesize$0.076$ \\
Episodic            & $\mathbf{100.0 \pm 0.0}$ & $\mathbf{96.3 \pm 3.1}$ & $\mathbf{90.0 \pm 1.7}$ & $43.0 \pm 2.6$ \\
                    & \footnotesize$\mathbf{0.849}$ & \footnotesize$\mathbf{0.701}$ & \footnotesize$\mathbf{0.605}$ & \footnotesize$0.117$ \\
Summary             & $43.0 \pm 2.7$ & $51.7 \pm 5.1$ & $46.7 \pm 1.5$ & $\mathbf{45.7 \pm 4.5}$ \\
                    & \footnotesize$0.136$ & \footnotesize$0.186$ & \footnotesize$0.149$ & \footnotesize$\mathbf{0.168}$ \\
Semantic            & $15.0 \pm 2.5$ & $10.0 \pm 3.0$ & $24.0 \pm 2.6$ & $25.0 \pm 5.3$ \\
                    & \footnotesize$0.050$ & \footnotesize$0.035$ & \footnotesize$0.056$ & \footnotesize$0.049$ \\
Hybrid              & $39.0 \pm 3.0$ & $40.3 \pm 3.2$ & $45.3 \pm 3.8$ & $45.3 \pm 0.6$ \\
                    & \footnotesize$0.128$ & \footnotesize$0.158$ & \footnotesize$0.132$ & \footnotesize$0.158$ \\
\bottomrule
\end{tabular*}
\tabnote{Test start fixed at the original pose; columns vary the demonstration used for memory. SR [\%]: mean $\pm$ std; SPL below: mean. Original-start results repeat Table~\ref{tab:main}; others use three runs. Semantic also varies in document length, limiting attribution to start pose alone.}
\end{table}

\textbf{Scene descriptions remain useful from a new start.} Summary is less sensitive to changes in the test start than Full-context and Episodic memory (Table~\ref{tab:main}). Its descriptions of item locations, landmarks, passages and routes can remain useful when the agent starts elsewhere: changing the starting pose does not change the scene itself. The agent still has to work out how to reach its goals, but it need not follow a demonstration from the beginning. Summary also has a lower original-start SR, which leaves less room for a decline. Summary still outperforms Working memory, but its advantage narrows by $5.8$ percentage points as the baseline gains $7.1$ points at the new test start; nearly constant SR therefore does not mean an unchanged benefit from memory. Semantic memory provides general spatial and navigation knowledge, such as approaching targets through accessible aisles or choosing detours, but does not consistently improve on Working memory.

\textbf{Historical actions need to fit the current episode.} Full-context and Episodic memory supply expert actions, either as retained trajectory records or a retrieved local continuation. Their high original-start SR is consistent with the value of this guidance when the current episode matches the demonstration. The action-versus-image comparison in Sec.~\ref{sec:design} supports this interpretation for Episodic memory. Reusing these actions from a different start requires relating the current state to the stored trajectory and finding where to join or how to adjust its route. This may help explain their greater sensitivity to start changes. The mismatch also affects successful execution: with opposite-heading and nearby-start demonstrations, SPL falls proportionally more than SR (Table~\ref{tab:transfer}), indicating lower average path efficiency among the episodes that succeed.

\textbf{Access strategy affects how well trajectory memory transfers.}
Episodic memory retains higher SR and SPL than Full-context memory with opposite-heading and nearby-start demonstrations, suggesting an advantage from matching the current view to a local continuation rather than leaving the policy to select from the full trajectory. This benefit weakens for distant-start demonstrations, showing that local retrieval alone does not guarantee successful reuse across larger pose mismatch. The same distinction appears when more experience is available: at the new test start, increasing from one to four demonstrations changes SR only modestly for Full-context, Summary, and Semantic memory, but improves Episodic memory by $14.3$ percentage points (Table~\ref{tab:main}). Hybrid does not exceed Summary in any reported comparison, indicating that simply concatenating Summary and Semantic memory provides no additional SR gain. These results suggest that preserving historical information is not sufficient; how that information is accessed and matched to the current episode also matters. We examine the effect of additional and mixed histories further in RQ3.

\subsection{RQ2: Acting with Outdated Memory}
\label{sec:rq2}

\begin{table}[t]
\centering
\caption{\textbf{RQ2: Responses to a blocked route.}}
\label{tab:rq2}
\footnotesize
\setlength{\tabcolsep}{3.5pt}
\begin{tabular*}{\columnwidth}{@{\extracolsep{\fill}}lccccc@{}}
\toprule
\textbf{Representation} & $\Delta$ & \emph{reten.} & \textbf{Follow [\%]} & \textbf{Recovery [\%]} \\
\midrule
Working memory      & $0$     & $0.45$ & $64.7 \pm 3.1$ & $11.8 \pm 4.8$ \\
Full-context        & $-7.7$  & $\mathbf{0.58}$ & $71.0 \pm 1.0$ & $\mathbf{30.5 \pm 2.9}$ \\
Summary             & $-14.3$ & $\mathbf{0.39}$ & $\mathbf{80.3 \pm 0.6}$ & $\mathbf{15.3 \pm 3.8}$ \\
Semantic            & $-0.3$  & $0.25$ & $59.0 \pm 2.0$ & $6.3 \pm 2.7$ \\
Hybrid              & --- & --- & --- & --- \\
Episodic            & $-16.7$ & $0.56$ & $63.3 \pm 2.1$ & $29.6 \pm 4.7$ \\
\bottomrule
\end{tabular*}
\tabnote{New test start and four demonstrations per task, as in the unchanged-route reference (Table~\ref{tab:main}). $\Delta$: SR change minus the working-memory change; retention: blocked/unchanged SR. Follow: percentage of episodes attempting a blocked cell; recovery: success among those episodes. Follow and recovery: mean $\pm$ std over three runs (Sec.~\ref{sec:metrics}).}
\end{table}

\textbf{Route changes with varied prior experience.} We use four demonstrations per task, aiming to give the agent more varied experience to draw on when a route becomes blocked. Within our evaluation budget, we prioritize this setting over an additional single-demonstration comparison. The unchanged and blocked route conditions use the same demonstration set and new test start. SR falls for every evaluated configuration, including Working memory (Table~\ref{tab:main}). Blocking thus makes navigation harder even without historical memory; for the other configurations, it also leaves part of their route guidance outdated. Table~\ref{tab:rq2} examines these outcomes through baseline-relative change, retention, obstacle encounters and subsequent task completion.

\textbf{Start-pose tolerance does not imply tolerance to outdated routes.} Summary connects RQ1 and RQ2: changing the start leaves its scene descriptions valid, whereas blocking a passage invalidates part of its route knowledge. Item locations and landmarks may still be useful, but reaching them can require a different path. Summary's high follow rate and low conditional recovery suggest that the ability to reuse scene descriptions across starts does not extend equally to completing tasks after a route change. These measures capture encounters and eventual completion, rather than repeated attempts or the policy's interpretation of feedback.

\textbf{Useful action guidance does not guarantee recovery.} Full-context and Episodic memory retain more of their unchanged-route success than Summary and still outperform Working memory after blocking. Historical trajectories therefore remain useful in this setting despite containing an invalid route segment. Yet their conditional recovery rates are both around $30\%$: Episodic memory's higher overall blocked-route SR does not establish better recovery than Full-context memory. Their remaining advantage in task completion should be distinguished from the ability to adapt an ongoing plan after encountering an obstruction. Semantic memory's small baseline-relative change also needs care: its low absolute success and retention provide little evidence of robustness to the route change.

The results motivate separating still-valid target and location knowledge from outdated route information, then using feedback to revise the latter. This is a direction for improving adaptation; the present protocol evaluates decisions with fixed historical memory and does not test selective memory updating.

\subsection{RQ3: Using Multiple Histories}
\label{sec:rq3}

\begin{figure}[t]
\centering
\captionsetup{font=footnotesize}
\includegraphics[width=0.75\columnwidth]{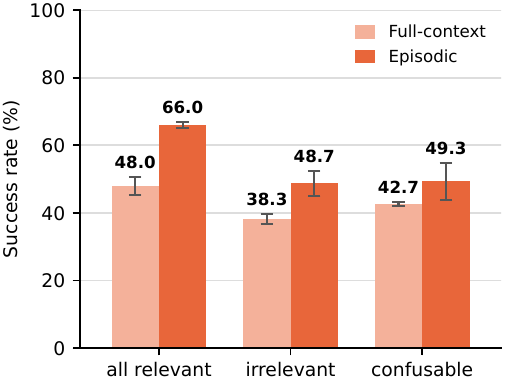}
\caption{\textbf{RQ3: Success under mixed histories.}
New start, unchanged route, and four demonstrations per task; mixed conditions replace two with other-task histories (Sec.~\ref{sec:variants}). Mean SR $\pm$ std over three 100-case runs.}
\label{fig:rq3}
\end{figure}

\textbf{Making use of additional experience.} Increasing from one to four relevant demonstrations per task improves Episodic SR by $14.3$ percentage points, while the other representations show only small changes (Table~\ref{tab:main}). As in RQ1, Episodic retrieves a local continuation for the current view; additional demonstrations expand its candidate pool without increasing the number of segments supplied per decision. Full-context shares a fixed image budget across more trajectories; Summary and Semantic supply larger documents without selecting content for the current state. With four demonstrations per task, Summary uses fewer images per request than Episodic ($7$ versus $14$) but more inference tokens per 100 cases ($588$M versus $35$M), without a mean SR gain over one demonstration. RQ1 and RQ3 suggest that the usefulness of experience depends on both its content and how the policy accesses it at decision time.

\textbf{Mixing histories from different tasks.} We next keep four demonstrations per task but replace two with histories from other tasks with different target requirements. We compare histories with no shared poses along the reference route for the new test start (\emph{irrelevant}) against histories that share poses and are selected for potentially conflicting continuations (\emph{confusable}). Both replacements reduce success for Episodic and Full-context memory (Fig.~\ref{fig:rq3}). Performance therefore depends on the composition of the demonstration set even when its size is fixed. Because replacement removes relevant experience as well as introducing other tasks' histories, the observed decrease cannot be attributed entirely to interference.

\textbf{Retrieval is only one step in using memory.} Episodic retrieval selects other tasks' histories more often in the confusable condition ($66.9\%$, two runs) than in the irrelevant condition ($45.5\pm1.5\%$, three runs), yet mean SR is not lower. One possible explanation is that a trajectory collected for a different goal can still contain useful local navigation; the policy also receives the current instruction and observation when deciding how to use a retrieved segment. A history's task of origin therefore need not determine the value of every part of it. The logs establish which history was retrieved, but do not establish its effect on the subsequent action. Their different trajectory lengths limit interpretation. Similar mean SR does not establish equivalent performance or show that either replacement is harmless. 

\subsection{Task Requirements and Memory Benefits}
\label{sec:tasktype}



\begin{table}[t]
\centering
\caption{\textbf{Task completion across navigation requirements.}}
\label{tab:tasks}
\footnotesize
\setlength{\tabcolsep}{3.5pt}
\begin{tabular*}{\columnwidth}{@{\extracolsep{\fill}}lcccc@{}}
\toprule
\textbf{Representation} & \textbf{Basic} & \textbf{Multi-object} & \textbf{Loop} & \textbf{T9} \\
\midrule
Working memory & $53$ / $30$ & $22$ / $9$ & $10$ / $1$ & $0$ / $0$ \\
Full-context   & $85$ / $72$ & $60$ / $28$ & $18$ / $7$ & $17$ / $3$ \\
Summary        & $72$ / $35$ & $55$ / $25$ & $19$ / $1$ & $20$ / $0$ \\
Semantic       & $38$ / $10$ & $18$ / $4$ & $8$ / $1$ & $0$ / $3$ \\
Episodic       & $\mathbf{93}$ / $\mathbf{67}$ & $\mathbf{78}$ / $\mathbf{45}$ & $\mathbf{22}$ / $\mathbf{11}$ & $\mathbf{93}$ / $\mathbf{23}$ \\
\bottomrule
\end{tabular*}
\tabnote{New test start and four demonstrations per task. Each cell shows mean SR [\%] over three runs for unchanged / blocked routes. Groups follow Table~\ref{tab:task_taxonomy}: basic search (T1--T2), multi-object search (T3--T6), and return-to-origin loops (T7, T8, T10). T9, which revisits the same object instance, is shown separately.}
\end{table}

Table~\ref{tab:tasks} shows that memory benefits depend strongly on task requirements. Full-context, Summary and Episodic outperform Working memory on basic and multi-object search under both unchanged and blocked routes, whereas Semantic remains below Working memory in both task groups under both route conditions. Historical scene and trajectory information therefore helps locate targets and complete visits across multiple locations, although the benefit varies by representation.

Return tasks expose a different limitation. T7, T8 and T10 require the agent to revisit the current episode's origin after completing intermediate goals, and all representations achieve relatively low success, suggesting that historical experience alone does not reliably support this form of return. T9 instead requires revisiting the same object instance after visiting another object. Here, Episodic memory shows a clear advantage, consistent with the value of retrieving local historical actions and views for instance-specific revisiting. This advantage decreases when the route is blocked but remains the strongest among the evaluated representations. The task-level scores do not isolate instance identification from failures at other execution stages.

The contrast between T9 and the return-to-origin tasks helps clarify what the stored trajectories support. Under unchanged routes, Episodic memory reaches 93\% on T9, versus at most 20\% for the other evaluated representations. On T7, T8 and T10, it reaches 22\% against Working memory's 10\%, its smallest advantage across the four task groups under this condition. T9 revisits a target instance also visited in the demonstrations, whereas the return tasks require reaching the current episode's own origin. This pattern suggests that view-matched retrieval is more useful for revisiting demonstrated targets than for returning to an episode-specific starting location.

\subsection{Design Choices}
\label{sec:design}

\textbf{Common policy and settings.} Fixing the backbone, decoding settings and interaction interface lets us compare memory representations under a common decision policy. We fix Working memory at $h=6$, where SR peaks at $15.2\%$ in five-run original-start ablations, compared with $0.8\%$ without memory and $11.8\%$ at $h=20$. Episodic retrieval achieves $100\%$ with retrieved actions and $49.0 \pm 3.8\%$ with images alone. We use $H_{\text{epi}}=6$ with actions and resulting views, and cap Full-context reference images at $40$. These settings remain fixed across the three RQs.

\textbf{Original-start reference.} The first view matches an indexed expert frame. This is a favorable reference for reusing demonstrated routes; the RQ conditions test whether memory remains useful beyond it. The original-start prompt also states that the demonstration starts at the agent's current pose; removing this hint changed Full-context SR by two percentage points in a one-run-per-condition check.

\section{Limitations}
\label{sec:limitations}

Our study prioritizes controlled comparison over breadth, using one frozen VLM backbone, six representative memory formulations, and a standardized warehouse environment to isolate how memory affects embodied decision-making. The current evaluation focuses on navigation within a shared scene, with controlled variations in start pose, route availability, and the amount and relevance of prior experience. Future work could extend the framework to additional backbones, memory architectures, environments, and task domains, as well as larger and continually growing memory banks with more diverse or conflicting histories. Hybrid’s four-demonstration settings were not evaluated due to serving constraints. The historical memories considered here also remain fixed during execution; incorporating online verification, updating, and selective forgetting would enable agents to better handle outdated experience. Beyond navigation, the same evaluation principle can be applied to manipulation and other embodied tasks, providing a broader basis for studying how memory representation, composition, and access influence embodied decision-making.



\section{CONCLUSION}
\label{sec:conclusion}

We presented MemTransfer to compare six memory representations through three research questions about reusing past experience. Within the tested setting, high success at the original start can be accompanied by large declines when conditions change. Summary's smaller variation across starts does not extend to outdated routes: it retains less of its unchanged-route success than Working memory and the two trajectory representations. Additional relevant demonstrations benefit Episodic memory, whereas replacing relevant demonstrations with other-task histories reduces success for both Episodic and Full-context. These differences show why the conditions in which memory is used need to be considered when assessing its value.

These observations motivate closer attention to how an agent determines which parts of its experience apply to the current task. Selecting useful historical guidance and distinguishing valid knowledge from outdated route information are promising directions for improving memory use. Extending the comparison beyond one policy and warehouse would help establish which observations carry over to other embodied settings.

\section*{Acknowledgments}
The authors used generative AI tools during manuscript preparation to assist with generating and revising selected text and figures, as well as code generation and debugging.

This work was supported by the National Research Foundation of Korea (NRF) grant funded by the Korean government (MSIT) under the project ``Development of Risk-Enhanced Continual Learning for Embodied Intelligence in Long-Tail Environments'' (Grant No. RS-2026-25595451) and by the Korea Institute of Science and Technology Information (KISTI) R\&D program through the joint research project ``Development of the Next-Generation Integrated Wired/Wireless Communication Gateway (X-Gateway).''

\bibliographystyle{IEEEtran}
\bibliography{references}

\end{document}